\documentclass[conference]{IEEEtran}

\usepackage{amsfonts}
\usepackage{amssymb}
\usepackage{hyperref}
\usepackage{graphicx}
\usepackage{enumerate}
\usepackage{amsmath}
\usepackage{color}
\usepackage{amsthm}
\usepackage{verbatim}
\usepackage{amsmath}
\usepackage{algorithm}
\usepackage{cite}
\usepackage{algpseudocode}
\usepackage{subfig}
\usepackage{graphicx}
\usepackage{float}
\usepackage{autobreak}
\usepackage{amsfonts,amssymb}
\usepackage[fontsize=10pt]{fontsize}
\providecommand{\keywords}[1]{\textbf{\textit{Index terms---}} #1}
\newcommand{\bp}{\begin{proof} \small }
\newcommand{\ep}{\end{proof} \normalsize}
\newcommand{\epx}{\end{proof} \small}
\newcommand{\bpa}{\begin{proofappx} \footnotesize }
\newcommand{\epa}{\end{proofappx} \small }
\newtheorem{theorem}{Theorem}

\newtheorem{corollary}{Corollary}
\newtheorem{lemma}{Lemma}
\newtheorem{assumption}{Assumption}

\newtheorem*{theorem*}{Theorem}
\newtheorem*{proposition*}{Proposition}
\newtheorem*{corollary*}{Corollary}
\newtheorem*{lemma*}{Lemma}
\newtheorem*{assumption*}{Assumption}
\newtheorem*{definition*}{Definition}
\newtheorem*{claim*}{Claim}

\newcommand{\be}{\begin{equation}}
\newcommand{\ee}{\end{equation}}
\newcommand{\bs}{\begin{subequations}}
\newcommand{\es}{\end{subequations}}
\newcommand{\bq}{\begin{eqnarray}}
\newcommand{\eq}{\end{eqnarray}}
\newcommand{\bqn}{\begin{eqnarray*}}
\newcommand{\eqn}{\end{eqnarray*}}

\newcommand{\ba}{\left[ \begin{array}}
\newcommand{\ea}{\\ \end{array} \right]}
\newcommand{\ben}{\begin{enumerate}}
\newcommand{\een}{\end{enumerate}}
\def\real{{\mathchoice%
{\hbox{\rm\setbox1=\hbox{I}\copy1\kern-.45\wd1 R}}
{\hbox{\rm\setbox1=\hbox{I}\copy1\kern-.45\wd1 R}}
{\hbox{\scriptsize\rm\setbox1=\hbox{I}\copy1\kern-.45\wd1 R}}
{\hbox{\scriptsize\rm\setbox1=\hbox{I}\copy1\kern-.45\wd1 R}}}}

\def\Zint{{\mathchoice{\setbox1=\hbox{\sf Z}\copy1\kern-.75\wd1\box1}
{\setbox1=\hbox{\sf Z}\copy1\kern-.75\wd1\box1}
{\setbox1=\hbox{\scriptsize\sf Z}\copy1\kern-.75\wd1\box1}
{\setbox1=\hbox{\scriptsize\sf Z}\copy1\kern-.75\wd1\box1}}}
\newcommand{\complex}{ \hbox{\rm C\kern-0.45em\rule[.07em]{.02em}{.58em}%
\kern 0.43em}}

\allowdisplaybreaks
\begin{document}
\pagestyle{empty}

\title{Computation and Communication Efficient Lightweighting Vertical Federated Learning for Smart Building IoT}

\author{\IEEEauthorblockN{Heqiang Wang, Xiang Liu, Yucheng Liu, Jia Zhou, Weihong Yang, Xiaoxiong Zhong} \\
\IEEEauthorblockA{Department of New Network,\\
Peng Cheng Laboratory, Shenzhen, 518066, China}}

\maketitle

\begin{abstract}
With the increasing number and enhanced capabilities of IoT devices in smart buildings, these devices are evolving beyond basic data collection and control to actively participate in deep learning tasks. Federated Learning (FL), as a decentralized learning paradigm, is well-suited for such scenarios. However, the limited computational and communication resources of IoT devices present significant challenges. While existing research has extensively explored efficiency improvements in Horizontal FL, these techniques cannot be directly applied to Vertical FL due to fundamental differences in data partitioning and model structure. To address this gap, we propose a Lightweight Vertical Federated Learning (LVFL) framework that jointly optimizes computational and communication efficiency. Our approach introduces two distinct lightweighting strategies: one for reducing the complexity of the feature model to improve local computation, and another for compressing feature embeddings to reduce communication overhead. Furthermore, we derive a convergence bound for the proposed LVFL algorithm that explicitly incorporates both computation and communication lightweighting ratios. Experimental results on an image classification task demonstrate that LVFL effectively mitigates resource demands while maintaining competitive learning performance.
\end{abstract}

%\begin{IEEEkeywords}
%IEEEtran, journal, \LaTeX, paper, template.
%\end{IEEEkeywords}

% \fi
\IEEEpeerreviewmaketitle

\section{Introduction}
\begin{figure}[htp]
\vspace{-5pt}
\centering
\subfloat{\includegraphics[width=0.8\linewidth]{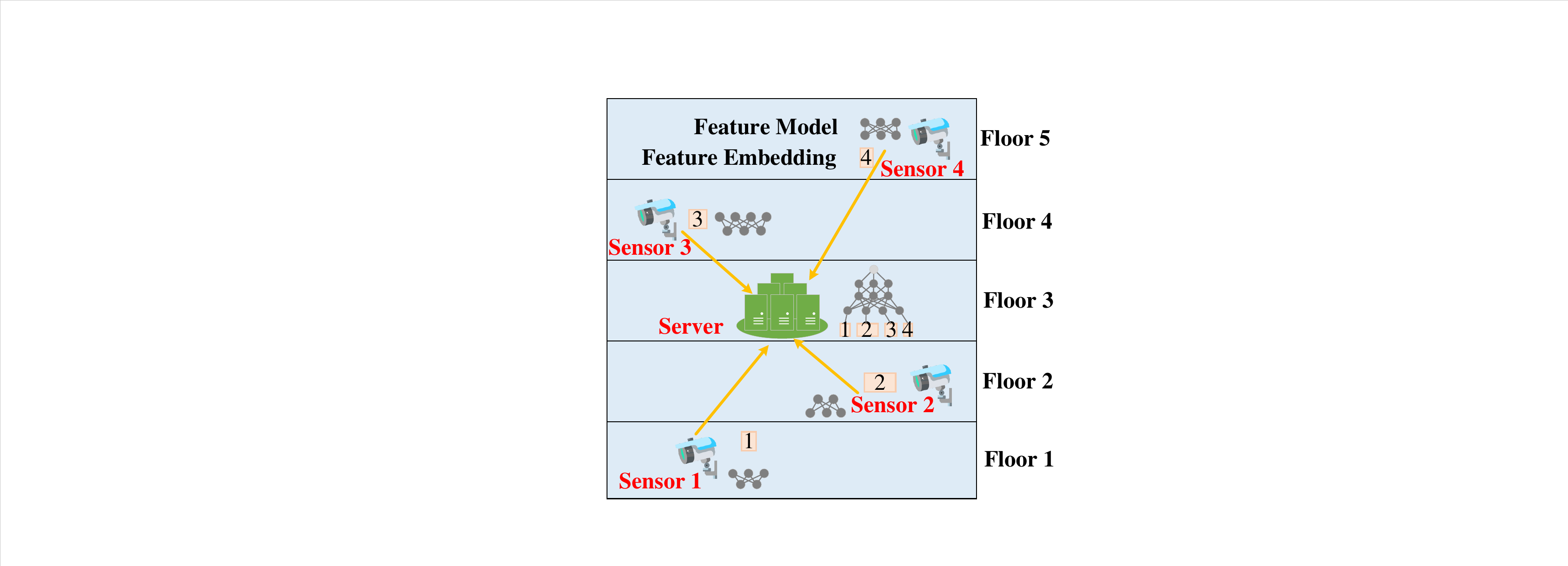}} 
\caption{Smart Building IoT based VFL}  
\label{VPFL}
\vspace{-10pt}
\end{figure}

As IoT devices in smart buildings continue to proliferate and improve in capability, their roles are expanding from basic sensing and control tasks to active participation in deep learning processes. Federated Learning (FL), a decentralized machine learning paradigm, enables multiple clients with locally stored data to collaboratively train a global model under the orchestration of a central server. By keeping data on edge devices, FL minimizes the need for data transmission and mitigates privacy risks. A typical application of FL in smart building environments is illustrated in Fig.~\ref{VPFL}, where sensor data collected from different floors represents distinct feature spaces, aligning naturally with the principles of Vertical Federated Learning (VFL). In VFL, each client holds a unique set of features corresponding to the same set of samples. During VFL training, each client develops a feature model, converting raw data features into a vector representation known as `feature embedding'. Instead of transmitting their feature models, clients in VFL send these feature embeddings to the server. The server then integrates these embeddings into a head model to determine the final loss.  This workflow highlights the fundamental differences between VFL and conventional FL, introducing a distinct set of challenges many of which cannot be effectively resolved by directly applying techniques designed for conventional FL. Consequently, addressing the unique requirements of VFL necessitates specialized solutions tailored to its structural.

During VFL training, the diversity among sensors results in variations in computational and communication capabilities, raising the need for synchronization solutions. While previous studies within conventional FL have addressed this, in conventional FL, local and global models are uniform due to identical feature spaces across clients, necessitating only local model pruning to meet specific client requirements. Conversely, in VFL, feature spaces differ among clients, leading to individual training on local feature models and subsequent uploading of trained feature embeddings to the server. This approach imposes distinct computational and communication demands for each sensor. Although prior research in VFL has examined feature embedding compression to reduce communication load  \cite{castiglia2022compressed, wang2023online}, the critical issue of computational burden has been largely neglected. Addressing this computational load is arguably a more pressing challenge for VFL deployment.

To tackle the diverse computational burdens encountered in the deployment of VFL in smart building IoT scenario, this paper introduces the concept of Lightweight Vertical Federated Learning (LVFL), designed to mitigate challenges in both computational and communication efficiency. The principal contributions of this study are summarized as follows: (1) This study introduces the LVFL framework, tailored to accommodate the varied communication and computation capabilities of heterogeneous workers. LVFL dynamically adjusts computational expenses for training feature models and communication costs for updating feature embeddings, ensuring efficiency across diverse system architectures.
(2) We establish the comprehensive convergence analysis of the LVFL algorithm and present the convergence bound that elucidates the relationship between the cumulative feature model and feature embedding lightweighting error, as well as the communication and computation lightweighting ratio. (3) Our experiments, conducted on the CIFAR-10 dataset, validate the capability of the LVFL algorithm to adjust the computation and communication lightweighting ratio both constantly and dynamically. The proposed LVFL algorithm yields comparable test accuracy levels while significantly reduced communication and computation burden. The theoretical proof details are accessible in the online supplementary material with link: https://github.com/ystex/LVFL/tree/main.

\section{Related Work}
In recent years, VFL has attracted significant attention. The concept of FL with vertically partitioned data was introduced in  \cite{hardy2017private}. Comprehensive surveys on VFL have been presented in \cite{yang2023survey, wei2022vertical, liu2022vertical}. However, VFL, distinct from conventional FL, poses its own set of challenges. Some research, such as \cite{feng2022vertical}, aims to optimize data utilization to enhance the joint model's effectiveness in VFL. In contrast, studies like \cite{ sun2021defending} focus on implementing privacy-preserving protocols to counter potential data leakage threats. Beyond these challenges, there has been exploration into improving training efficiency in VFL. These endeavors primarily target reducing communication overhead, either by allowing participants to conduct multiple local updates in each iteration \cite{liu2022fedbcd} or by compressing the data exchanged between parties \cite{li2020efficient}. While these studies primarily emphasize reducing communication overhead, a gap in these studies is the oversight of computational efficiency as a means to enhance training efficiency. 

Modern DNNs often comprise tens of millions of parameters \cite{simonyan2014very}. Storing and training such highly parameterized models on resource-constrained devices within FL can be challenging. Therefore, the lightweighting of models during training is an indispensable topic \cite{wei2021lightweight}. Considering the diverse computational capabilities of training devices, a viable strategy involves dynamically adjusting the size of the local model and diminishing its parameters, primarily via model pruning. Generally, model pruning techniques fall into two categories: unstructured pruning \cite{dong2017learning, sanh2020movement} and structured pruning \cite{ding2019centripetal, li2016pruning, you2019gate}. Unstructured pruning eliminates non-essential weights, specifically the inter-neuronal connections, in neural networks, resulting in significant parameter sparsity \cite{han2015learning}. However, the ensuing sparse matrix's irregularity poses challenges in parameter compression in memory, requiring specialized hardware or software libraries for efficient training \cite{han2016eie}. In contrast, structured pruning aims to discard redundant model structures, like convolutional filters \cite{li2016pruning}, without introducing sparsity. Consequently, the derived model can be considered as a subset or sub-configuration of the initial neural network, encompassing fewer parameters, thereby facilitating a reduction in computation overhead.

While a substantial portion of model pruning research centers on centralized learning contexts, recent studies have ventured into its application within FL settings \cite{jiang2022model, jiang2023computation}. The work in \cite{jiang2022model} presents PruneFL, an adaptive FL strategy that commences with initial pruning at a selected client and continues with subsequent pruning throughout the FL process, aiming to reduce both communication and computational overheads. The research outlined in \cite{jiang2023computation} introduces FedMP, a framework that employs adaptive pruning and recovery techniques to enhance both communication and computation efficiency, leveraging a multi-armed bandit algorithm for the selection of pruning ratios. Notably, the aforementioned model pruning techniques predominantly originate from conventional FL scenarios. However, due to the inherent differences between conventional FL and VFL, these techniques cannot be seamlessly applied to the VFL setting. Therefore driving the motivation for our study on model and feature embedding pruning in the VFL scenarios.

\section{System Model}
To begin, we define several foundational concepts of the VFL in smart building IoT. The VFL system consists of $K$ clients and a central server. The dataset, denoted as $\textbf{x} \in \mathbb{R}^{N \times D}$, has $N$ representing the total data samples and $D$ indicating the number of features. The row indexed by $i$ in $\textbf{x}$ corresponds to a data sample $x^i$. Each sample, $x^i$, possesses a unique set of features, $x^i_k$ retained by client $k$, and identified as a distinct subset. Every instance $x^i$ is paired with a label $y^i$. The vector $\textbf{y} \in \mathbb{R}^{N \times 1}$ represents all sample labels. Client $k$ maintains a dataset of local features, $\textbf{x}_k \in \mathbb{R}^{N \times D_k}$, with $D_k$ denoting the number of features for client $k$ and $i$-th row signifying the respective features $x^i_k$. In this context, we assume that both the server and all clients retain a copy of the labels $\textbf{y}$, consistent with previous studies.

Each client, represented by $k \in K$  has a unique set of feature model parameters, $\theta_k$. The server maintains the head model, $\theta_0$.  The function $h_k(\theta_k, x_k^i)$ denotes the feature embedding extracted from sample $x_k^i$ by client $k$. This feature embedding operation transforms the high-dimensional raw data into a lower-dimensional representation through multiple layers of DNN, effectively capturing essential information from the input data while significantly reducing its dimension. The overall model's parameters are collectively denoted as $\Theta = [\theta^\top_0, \theta^\top_1, ..., \theta^\top_K]^\top$. We can then express the learning objective as minimizing the subsequent equation:
\begin{align}
   F (\Theta; \textbf{x}; \textbf{y}) := \frac{1}{N} \sum_{i = 1}^N l(\theta_0, \{h_k(\theta_k; x^{i}_k)\}_{k=1}^K; y^i) 
\end{align}
where $l(\cdot)$ denotes the loss function for a single data sample. Within the server model framework, the equation $h_0(\theta_0, x^i) = \theta_0$ consistently applies. To enhance clarity and ensure consistency in notation throughout this paper, we implement several simplifications in our subsequent discussions. Firstly, the feature embedding for dataset $\textbf{x}_k$, is represented by $h_k(\theta; \textbf{x}_k)$, which we frequently abbreviate to $h_k(\theta_k; \textbf{x}_k) = h_k(\textbf{x}_k)$. Secondly, we allocate $k = 0$ to denote the head model in FC, where $h_0(\theta_0) = \theta_0$ for convenience. Finally, we often represent $F(\Theta) = F(h_0(\theta_0), h_1(\theta_1), ..., h_K(\theta_K))$, particularly in the subsequent algorithm details and convergence analysis. In the following section, we will elaborate on the specific algorithmic workflow of LVFL.

\section{Lightweighting Vertical Federated Learning}
In this section, we introduce the detail of LVFL algorithm in smart building IoT scenario. Considering the unique structural attributes of VFL, where computational demands are determined by the model size and communication requirements correlate with feature embedding size, the lightweighting methods derived from conventional FL are not directly applicable to VFL. To tackle this limitation, we adopt a progressive structured pruning strategy tailored for each client's feature model and a unstructured pruning strategy tailored for each client's feature embedding. Those approach is guided by the distinct computational and communication capacities of each client. Through these pruning techniques, we ensure consistent synchronization among all clients and facilitate efficient updates to the server in smart building IoT scenario.

In this scenario, we introduce a dual lightweighting ratio mechanism to distinctly optimize computation and communication. We denote $\alpha_k^t$ as the computation lightweighting  ratio and $\beta_k^t$ as the communication lightweighting  ratio. A higher $\alpha_k^t$ implies reduced demand for CPU cycles when processing a data sample. Correspondingly, a larger $\beta_k^t$ signifies a reduced need for embedding size during uploads. 
Suppose that each client performs $E$ local iterations before transmitting the model to the server, and the training process spans over a total of $R$ rounds, resulting in a cumulative duration of $T = RE$ local iterations. Algorithm \ref{alg: lvfl} showcases the workflow of LVFL, integrating the dual lightweighting ratio mechanism.

During each global round, when $t \mid E = 0$, every client will decide the unstructured pruned feature embedding $\hat{h}_k^{t} (\hat{\theta}_k^{t}; \textbf{x}_k)$ based on the communication lightweighting ratio $\beta_k^t$ (Line 5), and then transmit this pruned feature embedding to the server for the global update (Line 6). Once the server has gathered all embeddings, it proceeds to get the model representation $\hat{\Phi}^{t_0}$. Subsequently, it distributes the model representation to all clients (Lines 8-9), and $t_0$ is the start of the most recent global round when embeddings were shared. After receiving the model representation, each client needs to prune the feature model $\hat{\theta}_k^t$ according to its computation lightweighting ratio $\alpha_k^t$ (Line 11). Following this, each client will engage in $E$ rounds of local iterations using both the pruned embeddings received during iteration $t_0$ and its own unpruned embedding $ h_k (\hat{\theta}_k^{t}; \textbf{x}_k)$. The collection of embeddings from other clients, excluding client $k$, is represented as $\hat{\Phi}^{t_0}_{-k}$, and all embeddings are collectively denoted as $\hat{\Phi}^{t}_k$. During each local iteration, client $k$ will update the feature model $\hat{\theta}_k$ using mini-batch SGD with step size $\eta^{t_0}$ (Lines 15-16).

\begin{algorithm}
    \caption{LVFL} \label{alg: lvfl}
    \begin{algorithmic}[1]
        \State \textbf{Initialize}: The initial local model $\theta_k^0$ for all clients $k$ and server model $\theta_0^0$.
        \For {$t = 1, 2, ..., T - 1$}
            \If {$t \mid E = 0$}
                \For {$k = 1, 2, ..., K$ in parallel}
                    \State Determine $ \hat{h}_k^{t} (\hat{\theta}_k^{t}; \textbf{x}_k)$ according to $\beta_k^t$
                    \State Send $\hat{h}_k^{t} (\hat{\theta}_k^{t}; \textbf{x}_k)$ to server
                \EndFor
            \State Server collects model representation $\hat{\Phi}^{t_0}$
            \State Server sends $\hat{\Phi}^{t_0}$ to all clients
                \For {$k = 1, 2, ..., K$ in parallel}
                    \State Prune the feature model $\hat{\theta}_k^t$ according to $\alpha_k^t$
                \EndFor
            \EndIf
            \For {$k = 0, 1, 2, ..., K$ in parallel}
                \State Obtain $\hat{\Phi}^{t}_k\leftarrow \left \{ \hat{\Phi}^{t_0}_{-k},   h_k (\hat{\theta}_k^{t}; \textbf{x}_k)\right \}$
                \State Update feature or head model $\hat{\theta}_k^{t+1}$.
            \EndFor
        \EndFor
    \end{algorithmic}
\end{algorithm}

We utilize an element-wise product to express the pruned feature model and feature embedding. Specifically, the pruned feature model can be represented as $\hat{\theta}_k^t = \theta_k^t \odot \textbf{m}_k^t$, where $\theta_k^t$ denotes the original structure of client $k$'s local model without pruning, and $\textbf{m}_k^t$ signifies the mask vector, containing zeros for parameters in $\theta_k^t$ that are pruned. 
Similarly, for the pruned feature embedding, we express it as $\hat{h}_k^t (\hat{\theta}_k^t) = h_k^t (\hat{\theta}_k^t) \odot \textbf{l}_k^t$, where $h_k^t (\hat{\theta}_k^t)$ represents the initial embedding of client $k$ without pruning, and $\textbf{l}_k^t$ denotes the mask vector with zeros for parameters in $h_k^t (\hat{\theta}_k^t)$ that have been pruned. During each global round, the client's feature model and feature embedding undergo adjustment based on the computation lightweighting ratio $\alpha_k^t$ and the communication lightweighting ratio $\beta_k^t$. Next we explain in detail how to prune feature model and feature embedding to achieve lightweight separately.

\textbf{Feature Model Lightweighting}. In the feature model, we adopt a structured model pruning method to adjust the feature model from ${\theta}_k^t$ to $\hat{\theta}_k^t$, guided by the computation lightweighting ratio $\alpha_k^t$. This approach is consistent with existing research. To simplify the model and avoid introducing layer-specific hyper-parameters, we apply a uniform pruning ratio across all layers, as suggested by previous studies \cite{li2016pruning}. Within each layer, filters or neurons are ranked based on their importance scores, and those with lower scores are pruned based on the predetermined lightweighting ratio. We recommend employing the $l_1$ norm to calculate these importance scores. 

\textbf{Feature Embedding Lightweighting}. For feature embedding, we employ the unstructured model pruning method to refine the feature embedding from $h_k^t (\hat{\theta}_k^t)$ to $\hat{h}_k^t (\hat{\theta}_k^t)$, guided by the communication lightweighting ratio $\beta_k^t$. This approach involves nullifying weights of the lowest absolute values to meet the pruning criteria. Although this method preserves the structure of the embedding, it substantially reduces communication costs by transmitting only the non-zero values.

Leveraging the aforementioned mechanism, clients can streamline both the feature model and feature embedding, directed by the parameters $\alpha_k^t$ and $\beta_k^t$. Our approach seeks to optimally diminish the model's complexity based on demand. Nevertheless, given that structured pruning is applied to the feature model, it is imperative to reassess the necessity for further pruning of the feature model prior to the training begin of each global round. Conversely, for feature embedding, which utilizes unstructured pruning, no additional verification step is required. Subsequently, we elucidate the mechanism for determining the computation lightweighting ratio of the feature model in detail.

\begin{enumerate}
    \item \textbf{If $\alpha_k^{t} > {\alpha}_k^{t-}$}. Where ${\alpha}_k^{t-} = \max \left \{{\alpha}_k^{1}, \cdots,  {\alpha}_k^{t-1} \right \}$ denotes the maximum computation lightweighting ratio previously attained by client $k$'s feature model before global round $t$. In this case further computation lightweighting is executed to meet the required lightweighting ratio for the current round.
    \item \textbf{If $\alpha_k^{t} \leq {\alpha}_k^{t-}$}.  Since the computation lightweighting ratio has been attained, the existing feature model was sufficiently streamlined, negating the need for further lightweighting in the current round.
\end{enumerate}
This iterative process is repeated until the training converges or until predetermined termination conditions are satisfied. The subsequent section will delve into the analysis of the convergence bound of the LVFL algorithm.

\section{Convergence Analysis}
In this section, we will delve into the convergence analysis of our LVFL algorithm. To begin, we need to establish some notations and definitions that will be utilized in the subsequent discussion. Specifically, we will define two types of errors that arise from the lightweighting mechanisms: communication lightweighting error and computation lightweighting error.

\textbf{Communication Lightweight Error}: This error quantifies the degree to which the lightweighting feature embedding is an accurate approximation of the original feature embedding. It is mathematically expressed as $\epsilon_k^t:= h_k^t ({\theta}_k^t) - \hat{h}_k^t ({\theta}_k^t)$. The squared communication lightweighting error from client $k$ at round $t$ is denoted as $\Omega_k^t \triangleq \mathbb{E}\| \epsilon_k^t  \|^2$.

\textbf{Computation Lightweight Error}:
This error assesses how closely the lightweighting feature model resembles the original feature model. It is defined as $\varphi_k^t:=   h_k^t({\theta}_k^t) -  h_k^t(\hat{\theta}_k^t)$. The squared computation lightweighting error from client $k$ at round $t$ as $\Psi_k^t \triangleq \mathbb{E}\| \varphi_k^t  \|^2$.

Let $\hat{\textbf{G}}^t$ be the stacked partial derivatives at iteration $t$:
\begin{align}
    \hat{\textbf{G}}^t := \left [ (\triangledown_0  F (\hat{\Phi}^{t}_0; \textbf{y}))^T, \dots, (\triangledown_K  F (\hat{\Phi}^{t}_K; \textbf{y}))^T \right ]^T
\end{align}
Then the global model updates as: $\Theta^{t+1} = \Theta^{t} - \eta^{t_0}  \hat{\textbf{G}}^t$.

We define ${\Phi}^{t_0}$ to be the set of embeddings that would be received by each client at iteration $t_0$ if no computation and communication lightweight error were applied: ${\Phi}^{t_0} \leftarrow \left \{ \theta_0^{t},   {h}_1^{t} ({\theta}_1^{t}), \cdots   {h}_k^{t} ({\theta}_k^{t})\right \}$. Here we define $ {\Phi}^{t_0}_{-k}$ be the set of embeddings from other parties $j \neq k$, and we have $ {\Phi}^{t_0}_{k} := \left \{ {\Phi}^{t_0}_{-k},   h_k ({\theta}_k^{t}; \textbf{x}_k)\right \}$. Our convergence analysis will utilize the following standard assumptions about the VFL. 
\begin{assumption}[Smoothness]
There exists positive constants $L < \infty $ and $L_k < \infty $, such that for all $\Theta_1, \Theta_2$, the objective function satisfies: $\left \| \triangledown F(\Theta_1) - \triangledown F(\Theta_2)   \right \| \leq L \left \| \Theta_1 - \Theta_2  \right \|$ 
and $\left \| \triangledown_k F(\Theta_1) - \triangledown_k F(\Theta_2)   \right \| \leq L_k \left \| \Theta_1 - \Theta_2  \right \|$.
\label{assm:SMO}
\end{assumption}
\begin{assumption}[Bounded Hessian] There exists positive constants $H_k$ for $k = 0, \dots, K$ such that for all $\Theta$, the
second partial derivatives of $F$ satisfy: $\left \| \triangledown^2_{h_k} F(\Theta)   \right \|_{\mathcal{F}}  \leq H_k$. Where $\left \| X \right \|_{\mathcal{F}}$ is the Frobenius norm of a matrix $X$.
\label{assm:BH}
\end{assumption}
\begin{assumption}[Bounded Embedding Gradients] There exists positive constants $G_k$ for $k = 0, \dots, K$ such that for all $\theta_k$, the embedding gradients are bounded: $\left \| \triangledown_{\theta_k}  h_k(\theta_k; x_k) \right \|_{\mathcal{F}}   \leq G_k$. 
\label{assm:BEG}
\end{assumption}

Assumption 1 guarantees that the function's slope changes smoothly without any sudden or drastic alterations. Assumption 2 controls the curvature of the function, preventing it from displaying extreme or erratic behavior. Assumption 3 manages the changes in the embedding, ensuring that they don't result in severe fluctuations, thereby aiding in the stabilization of the learning process. With those assumptions, we can get the following Lemma.

\begin{lemma}\label{lemma1}
The norm of the difference between the objective function value with computation and communication lightweighting and without computation and communication lightweighting is bounded as:
\begin{align}
   & \mathbb{E} \|\nabla_k F(\hat{\Phi}^{t}_k) - \nabla_k F({\Phi}^{t}_k)\|^2 \notag \\ 
   \leq & 2 G_k^2  H_k^2  \sum_{j\neq 0}^K  \Psi_k^{t_0}   
   + 2 G_k^2  H_k^2  \sum_{j\neq 0, j\neq k}^K  \Omega_k^{t_0}  
\end{align}
\end{lemma}
Using Lemma \ref{lemma1}, we can bound the effect of lightweighting errors in detail. Afterwards we can derive the Theorem \ref{thm:general}:
\begin{theorem}\label{thm:general}
Under Assumption 1-3, the average squared gradient over $R$ global rounds of the LVFL is bounded as:
\begin{align}
     & \frac{1}{R} \sum_{t_0 = 0}^{R-1}\mathbb{E}\|\nabla F(\Theta^{t_0})\|^2  \leq \frac{3\left [ F(\Theta^{0}) - \mathbb{E}[F(\Theta^{T})] \right ] }{\eta T} \notag \\
    & + \frac{108 E^2}{R} \sum_{t_0 = 0}^{R-1} \sum_{k=0}^K H_k^2 G_k^2 \sum_{j\neq 0}^K  \Psi_j^{t_0}  \notag \\
    &  + \frac{108 E^2 }{R} \sum_{t_0 = 0}^{R-1}  \sum_{k=0}^K G_k^2  H_k^2 \sum_{j\neq 0, j\neq k}^K  \Omega_j^{t_0} 
\end{align}
\end{theorem}
The proof can be found in Appendix of online supplementary materials. The convergence bound detailed above unveils several insights: The first term illustrates the difference between the initial and final models. The second term highlights the error stemming from the use of the lightweighting feature model, while the third term represents an error caused by lightweighting feature embedding. This analysis reveals that the primary elements influencing the bound are the errors related to communication lightweighting and computation lightweighting. In other words, a larger error will result in a greater bound in the end. To delve deeper into the connection between error and lightweighting ratios, we must turn to the subsequent additional assumptions. 

\begin{assumption}[Bounded Embedding and Model] 
Positive constants $\delta > 0$ and $\mu > 0$ exist such that for $k = 0, \dots, K$, the following conditions are met:
$\mathbb{E} \left \|   h_k(\theta_k; x_k)    \right \|^2  \leq \delta^2, \quad \mathbb{E} \left \|   \theta_k   \right \|^2  \leq \mu^2$.
\label{assm:BEG}
\end{assumption}

\begin{assumption}[Lipschitz Continuous]
There also exists positive constants and $M_k < \infty $, such that for the all $\theta_1$ and $\theta_2$ satisfies:
$\left \|h_k(\theta_1) -  h_k(\theta_2)   \right \| \leq M_k \left \| \theta_1 - \theta_2  \right \|$.
\label{assm:conti}
\end{assumption}
Assumption 4 and Assumption 5 are required to build the connection between the lightweighting ratios and the lightweighting error. With these two assumptions, we are able to derive the subsequent result, referred to as Corollary \ref{coro1}.
\begin{corollary}\label{coro1} 
Under Assumption 1-5, with the computation lightweighting ratio $\alpha_k^t$ and communication lightweighting ratio $\beta_k^t$, we can further obtain the following bound:
\begin{align}
  & \frac{1}{R} \sum_{t_0 = 0}^{R-1}\mathbb{E}\|\nabla F(\Theta^{t_0})\|^2 \leq \frac{3\left [ F(\Theta^{0}) - \mathbb{E}[F(\Theta^{T})] \right ] }{\eta T} \notag \\
   & + \frac{108 E^2 \mu^2 }{R} \sum_{t_0 = 0}^{R-1} \sum_{k=0}^K H_k^2 G_k^2 M_k^2 \sum_{j\neq 0}^K  \alpha_j^{t_0}  \notag \\
    &  + \frac{108 E^2 \delta^2}{R} \sum_{t_0 = 0}^{R-1}  \sum_{k=0}^K G_k^2  H_k^2 \sum_{j\neq 0, j\neq k}^K  \beta_j^{t_0} 
\end{align}
\end{corollary}
The detailed proof for this result is located in Appendix within the online supplementary materials. As derived from Corollary \ref{coro1}, a key observation to emphasize is that the communication lightweighting error is governed by the communication lightweighting ratio $\beta_k^{t}$, while the computation lightweighting error is influenced by the computation lightweighting ratio $\alpha_k^{t}$. As a result, higher lightweighting ratios tend to result in more lenient convergence bounds, whereas lower ratios lead to more stringent convergence bounds. 

\section{Experiment}
In this section, we carry out experiments to assess the performance of LVFL. Specifically, we explore the effectiveness of the algorithm we've proposed by utilizing the VFL-based datasets: CIFAR-10. Subsequently, we will provide detailed information about the datasets and the corresponding models used in the experiment.

\textbf{CIFAR-10}: CIFAR-10 is a dataset used for object classification in images. In a specific training setup, there are 4 parties involved, with each party responsible for a different quadrant of each image. The number of training data samples $N=10000$, and the batch size $b_s = 256$. Every party (client) utilizes VGG16 as feature model for training, while the server focuses on training a 3-layer fully connected networks (FCN) as head model.

In the experiment, the following approaches are used for performance comparison. \textbf{No Lightweighting (NL)}, where no computational or communication lightweighting mechanisms are implemented. \textbf{Computational Lightweighting Only  (PL)}, applying solely computational lightweighting. \textbf{Communication Lightweighting Only (ML)}, utilizing exclusively communication lightweighting. \textbf{Lightweighting (L)}, incorporating both computational and communication lightweighting strategies to enhance efficiency. Next we first show the performance comparison of the above approaches with dynamic computational and communication lightweighting ratios.

\subsection{Performance Comparison} 
\begin{figure}[h]
\centering
\vspace{-10pt}
\subfloat[Test Accuracy]{\includegraphics[width=0.49\linewidth]{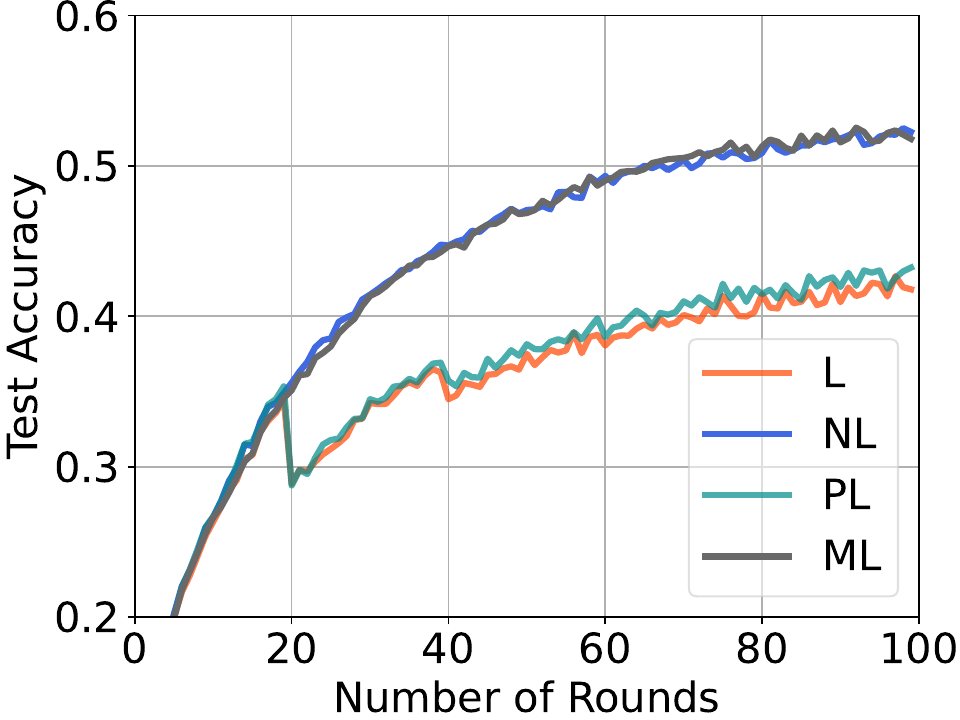}\label{lvfl-pc-compare}} 
\subfloat[Ratio By Rounds]{\includegraphics[width=0.49\linewidth]{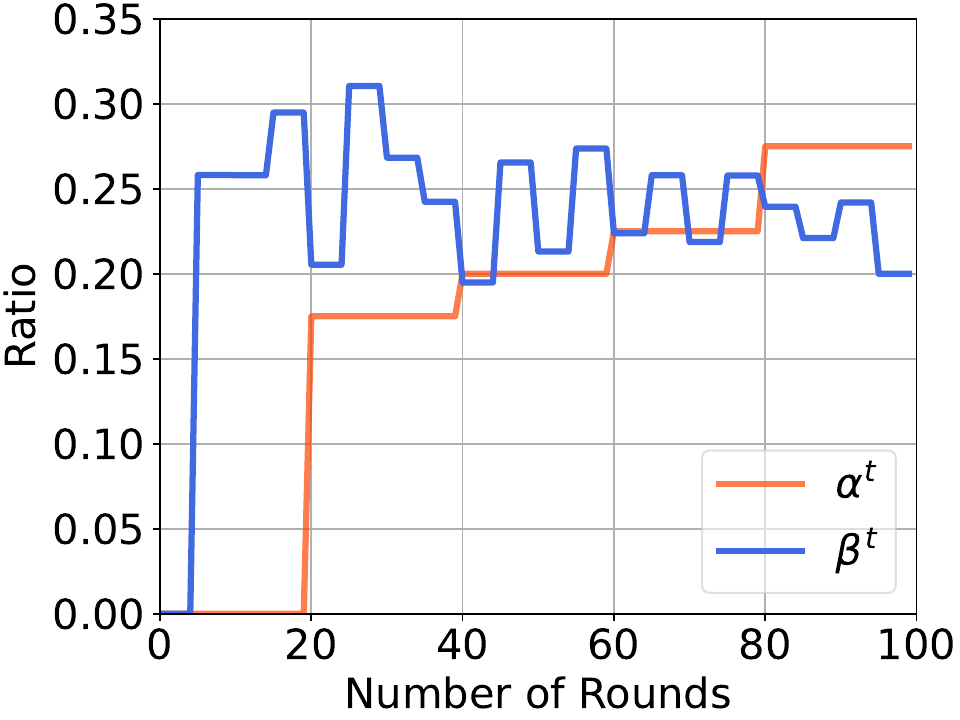}\label{lvfl-pc-ratio}} 
\caption{Performance Comparison By Round With CIFAR-10}   \label{lvfl-pc}
\vspace{-15pt}
\end{figure}

This section presents the learning performance optimized for computational and communication efficiency. At each round each client will be assigned the computational lightweighting ratio $\alpha_k^t$ and communication lightweighting ratio $\beta_k^t$. These ratios are then applied to individually tailor the lightweighting process to each client's unique requirements. For illustrative purposes, we assume a uniform set of ratios across all clients, with the specific values depicted in Fig.~\ref{lvfl-pc-ratio}. 
The learning performance associated with our approaches are illustrated in Fig.~\ref{lvfl-pc-compare}, from which several key observations can be derived. Primarily, computational lightweighting exerts a more significant influence on learning performance compared to communication lightweighting. Notably, computational lightweighting results in a marked decrease in test accuracy at the point of implementation, followed by a gradual recovery. Furthermore, the frequency and intensity of computational lightweighting are directly proportional to its impact on test accuracy.

\subsection{Effect of Ratio on Learning Performance} 
\begin{figure}[h]
\centering
\vspace{-15pt}
\subfloat[Computation Lightweighting ]{\includegraphics[width=0.49\linewidth]{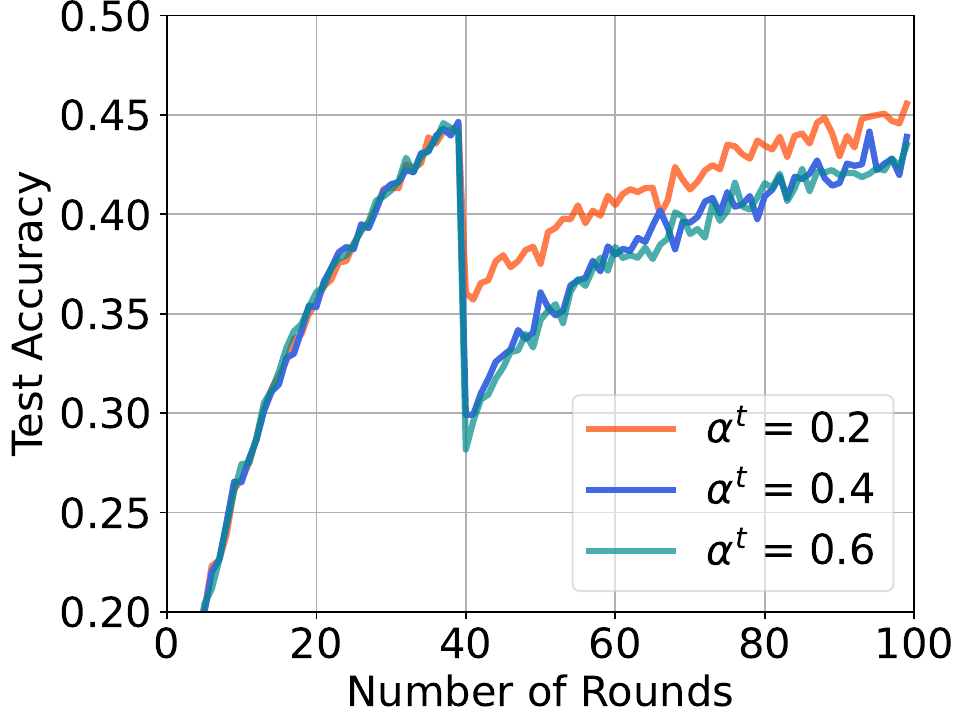}\label{lvfl-comp}} 
\subfloat[Communication Lightweighting ]{\includegraphics[width=0.49\linewidth]{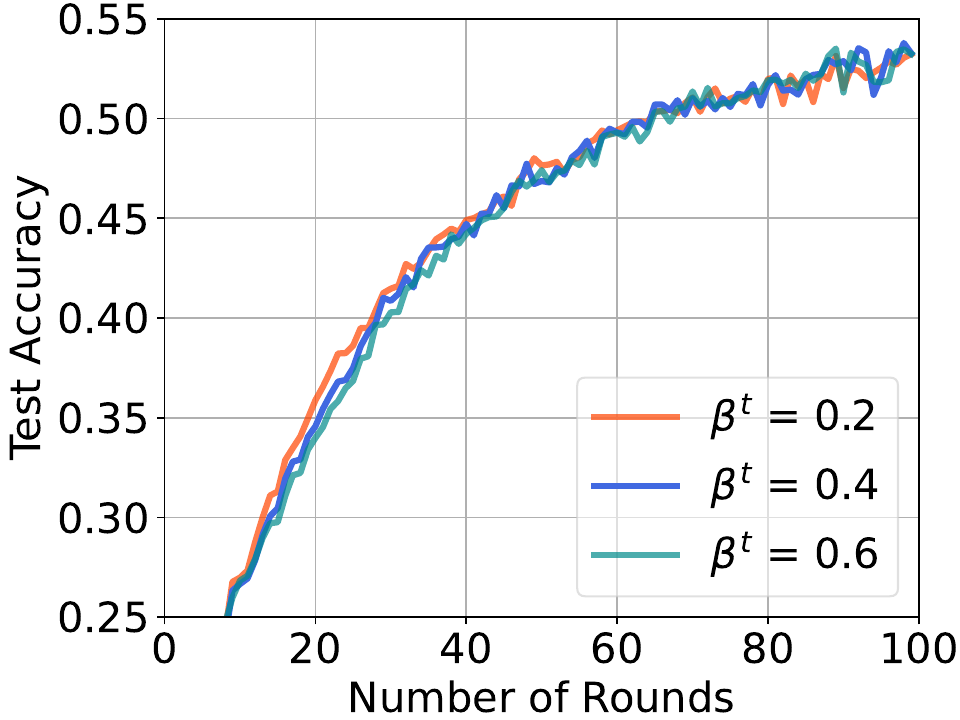}\label{lvfl-comm}} 
\caption{Effect of Ratio on Learning Performance} 
\vspace{-15pt}
\end{figure}

In this section we further investigate the effect of the choice of $\alpha_k^t$ and $\beta_k^t$ on learning performance. Here we explore the effects of computation and communication separately. In the scenario of computational lightweighting, as depicted in Fig.~\ref{lvfl-comp}, adjustments to the feature model are made at $t = 40$ with $\alpha^t \in [0.2, 0.4, 0.6]$. The outcomes indicate that higher $\alpha^t$ values result in more significant reductions in test accuracy, consequently necessitating a longer recovery period. In the context of communication lightweighting, as shown in Fig.~\ref{lvfl-comm},  varying communication lightweighting ratios $\beta^t \in [0.2, 0.4, 0.6]$ were applied to the feature embedding. The findings illustrate that lower $\beta^t$ values are associated with quicker convergence rates, particularly noticeable within the range $t \in [10, 40]$. However, the impact of communication lightweighting is not as substantial as that observed with computational lightweighting.

\section{Conclusion}
Our paper introduces the concept of Lightweight Vertical Federated Learning (LVFL) under smart building IoT scenario. Owing to structural distinctions between VFL and conventional FL, algorithm design and convergence analysis for VFL-based lightweight challenges significantly differ. Our convergence proofs elucidate the correlation between convergence bounds and the ratios of communication and computational lightweighting. Furthermore, the section on experimental results underscores the benefits of employing lightweight mechanisms. In the subsequent research, we intend to extend LVFL to be a more practical problem. We plan to formulate a long-term optimization problem that accurately represents the computational and communication difficulties clients face throughout the training phase. This will facilitate the determination of optimal lightweighting ratios specifically.

\bibliographystyle{IEEEtran}
\bibliography{bibligraphy}

\end{document}